\documentclass[10pt,twocolumn,letterpaper]{article}

\usepackage{cvpr}              

\usepackage{graphicx}
\usepackage{amsmath}
\usepackage{amssymb}
\usepackage{booktabs}
\usepackage{multirow}
\usepackage{makecell}
\usepackage{tabulary}

\usepackage[accsupp]{axessibility}

\usepackage[pagebackref,breaklinks,colorlinks]{hyperref}

\usepackage[capitalize]{cleveref}
\crefname{section}{Sec.}{Secs.}
\Crefname{section}{Section}{Sections}
\Crefname{table}{Table}{Tables}
\crefname{table}{Tab.}{Tabs.}

\def\cvprPaperID{9270} 
\def\confName{CVPR}
\def\confYear{2022}

\begin{document}

\title{Comprehending and Ordering Semantics for Image Captioning}

\author{Yehao Li, Yingwei Pan, Ting Yao, and Tao Mei \\
{\normalsize\centering JD Explore Academy}\\
{\tt\small \{yehaoli.sysu, panyw.ustc, tingyao.ustc\}@gmail.com, tmei@jd.com}
}

\maketitle

\begin{abstract}

Comprehending the rich semantics in an image and ordering them in linguistic order are essential to compose a visually-grounded and linguistically coherent description for image captioning. Modern techniques commonly capitalize on a pre-trained object detector/classifier to mine the semantics in an image, while leaving the inherent linguistic ordering of semantics under-exploited. In this paper, we propose a new recipe of Transformer-style structure, namely Comprehending and Ordering Semantics Networks (COS-Net), that novelly unifies an enriched semantic comprehending and a learnable semantic ordering processes into a single architecture.
Technically, we initially utilize a cross-modal retrieval model to search the relevant sentences of each image, and all words in the searched sentences are taken as primary semantic cues.
Next, a novel semantic comprehender is devised to filter out the irrelevant semantic words in primary semantic cues, and meanwhile infer the missing relevant semantic words visually grounded in the image. After that, we feed all the screened and enriched semantic words into a semantic ranker, which learns to allocate all semantic words in linguistic order as humans.
Such sequence of ordered semantic words are further integrated with visual tokens of images to trigger sentence generation.
Empirical evidences show that COS-Net clearly surpasses the state-of-the-art approaches on COCO and achieves to-date the best CIDEr score of 141.1\% on Karpathy test split.
Source code is available at \url{https://github.com/YehLi/xmodaler/tree/master/configs/image_caption/cosnet}.
\end{abstract}

\section{Introduction}
\label{sec:intro}

\begin{figure}[!tb]
\vspace{-0.39in}
         \centering\includegraphics[width=0.43\textwidth]{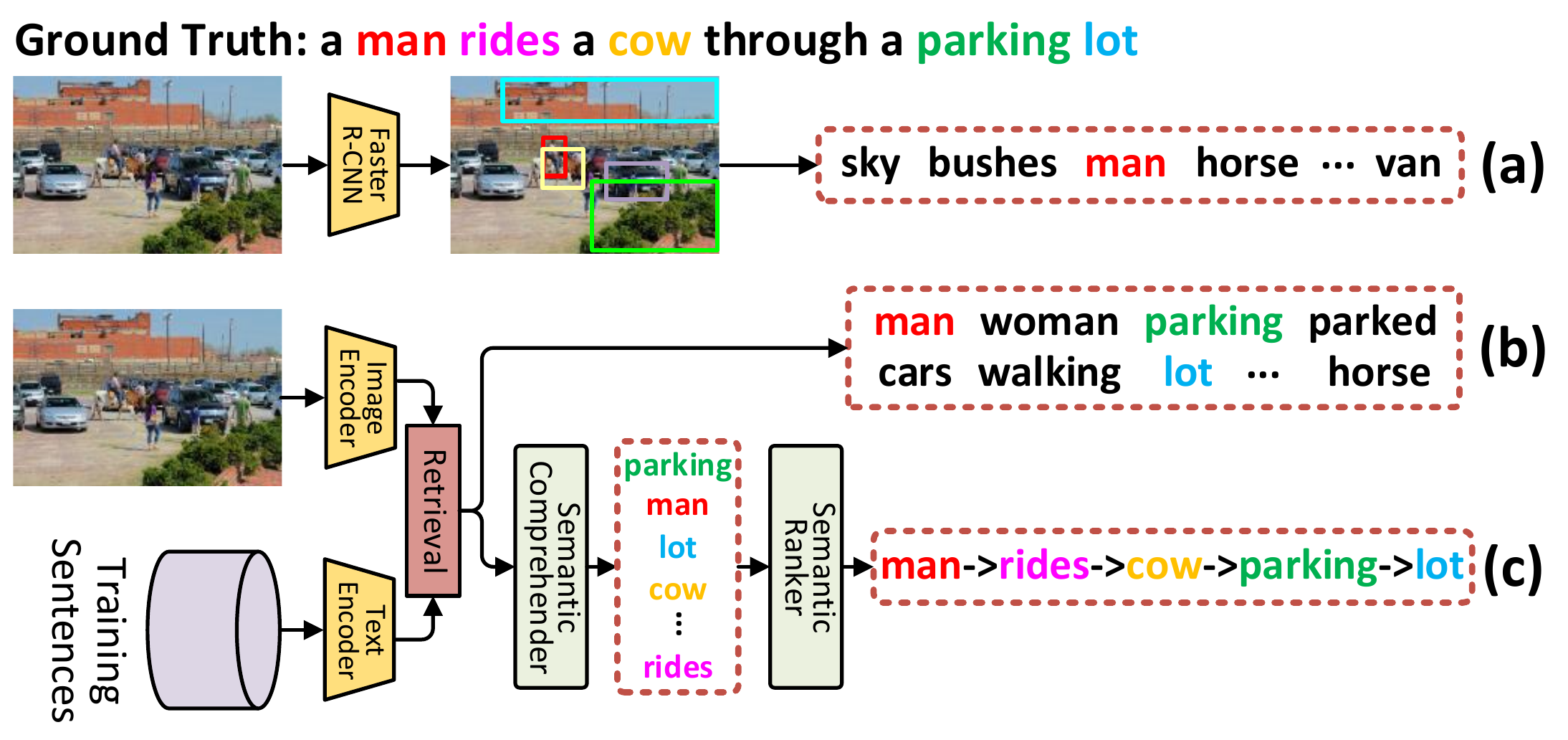}
         \vspace{-0.15in}
         \caption{\small Semantics produced by (a) pre-trained object detector, (b) cross-modal retrieval model (CLIP), and (c) our semantic comprehender \& ranker for image captioning.}
         \label{fig:intro}
   \vspace{-0.3in}
\end{figure}

The ability to describe visual content with a descriptive utterance is a fundamental human capability that children are taught from childhood. To formalize such unique capability, the task of image captioning \cite{Mao:NIPS14,Fang:CVPR15,Karpathy:CVPR15,Vinyals14} is developed to simulate the human-like interaction between vision and language. The ultimate target of this task is to produce a \textbf{visually-grounded} and \textbf{linguistically coherent} sentence, which covers most semantics in an image that are worthy of mention and meanwhile describes them in linguistic order. Modern image captioning techniques generally focus on the former aspect of enhancing vision-language alignment by first capturing fine-grained semantics (e.g., attributes \cite{You:CVPR16,yao2017boosting}, objects \cite{anderson2017bottom,li2019pointing,yao2017incorporating}, or scene graph \cite{Yang:CVPR19,yao2018exploring,yao2019hierarchy}) via pre-trained image encoder (object detector/classifier). Then, a series of innovations that employ visual attention over these fine-grained semantics \cite{cornia2020meshed,huang2019attentio} are present to strengthen vision-language interaction.
However, the capability of semantic comprehending in pre-trained detector/classifier is severely limited by the pre-defined semantic/class labels. In addition, the pre-trained detector/classifier is not optimized along with sentence decoding process, thereby hardly to be tuned for emphasizing visually salient semantics in output sentence. As shown in Figure \ref{fig:intro} (a), the pre-trained object detector (Faster R-CNN) solely captures one major semantic word (``man''), while the other mined semantic words are either irrelevant (e.g., ``horse'') or trivial (e.g., ``sky'' and ``bushes'').

To enhance the scalability and generalization of image encoder, a recent pioneering practice \cite{shen2021much} is to leverage CLIP model (i.e., image encoder and text encoder \cite{radford2021learning}) that is trained on diverse and large-scale data. In this work, we regard CLIP model as a powerful cross-modal retrieval model that retrieves relevant sentences from the human-annotated sentence pool. Such way naturally accumulates more salient semantic words that tend to be mentioned in visually similar images, while more irrelevant semantic words are also introduced (see Figure \ref{fig:intro} (b)). To alleviate this issue, we uniquely design a semantic comprehender that further refines the primary semantic cues in the searched sentences based on visual content. By doing so, the semantic comprehender (see Figure \ref{fig:intro} (c)) not only filters out the irrelevant semantic words (e.g., ``horse''), but also learns to infer the missing relevant semantic words (e.g., ``cow'' and ``rides''), pursuing an enriched and accurate semantic understanding.

In pursuit of the linguistical coherence of the output sentence, the recent advances directly capitalize on the RNN/Transformer based sentence decoder for language modeling. Unfortunately, such paradigm overly relies on the language priors, and sometimes leans to hallucinate semantic words that are not actually in an image, a phenomenon known as ``object hallucination'' \cite{rohrbach2018object}. Here we propose to mitigate the issue from the viewpoint of exploiting the inherent linguistic ordering of semantics as additional supervisory signals to guide sentence decoding process. Technically, a semantic ranker (see Figure \ref{fig:intro} (c)) is leveraged to rank all the refined semantic words derived from semantic comprehender in linguistic order, yielding a sequence of ordered semantic words. This semantic word sequence manifests the emphasis of the relative linguistic position of each semantic word in a sequence. As such, the sequence acts as the inherent skeleton of the descriptive sentence, and thus can be exploited to encourage the generation of relevant words at each decoding timestep.

In this work, we design a novel Transformer-style encoder-decoder structure for image captioning, namely Comprehending and Ordering Semantics Networks (COS-Net). Our launching point is to unify the above-mentioned two processes of semantic comprehending and ordering into a single scheme, so that both semantic comprehender and ranker can be jointly optimized to better suit the sentence decoding procedure. Specifically, we first take the off-the-shelf CLIP as cross-modal retrieval model to retrieve semantically similar sentences for the input image. All semantic words in searched sentences are initially regarded as the primary semantic cues. Next, based on the output grid features of image encoder in CLIP, a visual encoder is utilized to contextually encode each grid feature into visual token via self-attention. By taking the primary semantic cues and visual tokens as inputs, semantic comprehender filters out irrelevant semantic words in primary semantic cues and meanwhile reconstructs the missing relevant semantic words through cross-attention mechanism. After that, semantic ranker learns to allocate all the refined semantic words in a linguistic order by upgrading each semantic word with the encoding of its estimated linguistic position. Finally, both the visual tokens and the ordered semantic words are dynamically integrated via attention to auto-regressively decode the output sentence word-by-word.

The main contribution of this work is the proposal of jointly comprehending and ordering the semantics in an image to boost image captioning. This also leads to the elegant views of how to nicely capture the richer relevant semantics that are worthy of mention from visual content, and how to explore the inherent linguistic ordering of them to further facilitate sentence generation. Extensive experiments on COCO demonstrate the effectiveness of our COS-Net.

\section{Related Work}

\textbf{RNN-based Encoder-decoder Scheme.}
In the deep learning era, researchers in \cite{Bahdanau14,Sutskever:NIPS14} demonstrate that the using of RNN-based encoder-decoder significantly improves machine translation. Subsequently, this RNN-based encoder-decoder scheme becomes the de-facto recipe of modern image captioning techniques. In analogy to the RNN-based sequence modeling in machine translation, the earlier attempts \cite{Mao:NIPS14,Vinyals14} directly employ the basic RNN-based encoder-decoder scheme for the task of image captioning, by encoding visual content with CNN and decoding output description with RNN. Next, the basic RNN-based scheme is upgraded with visual attention mechanism \cite{Xu:ICML15,Xiong2016MetaMind} that learns to dynamically pinpoint the most relevant local patches to boost the word prediction at each decoding timestep. Meanwhile, semantic attention mechanism \cite{You:CVPR16} is incorporated into RNN-based encoder-decoder to selectively emphasize the most relevant semantic attributes for sentence generation. After that, bottom-up and top-down attention \cite{anderson2017bottom} enables attention measurement at object level, rather than the conventional visual attention over equally-sized local patches. Scene graph structure \cite{Yang:CVPR19} that depicts the fine-grained semantics in an image is further integrated into the RNN-based encoder-decoder, aiming to exploit the language inductive bias.

\textbf{Transformer-based Encoder-decoder Scheme.}
Sparked by the breakthroughs in NLP field via Transformer \cite{vaswani2017attention}, numerous modern image captioning approaches capitalizing on Transformer-based encoder-decoder scheme start to emerge. The central spirit of this scheme aims to strengthen the visual encodings and vision-language interaction with self-attention or cross-attention mechanism in Transformer. Take for instance, in \cite{sharma2018conceptual}, the primary Transformer structure in NLP is directly employed for image captioning task. The spatial relations among objects are additionally incorporated into Transformer-based encoder-decoder in \cite{herdade2019image}. Recently, a series of innovations have been proposed to upgrade the attention mechanism in Transformer-style structure with attention gate \cite{huang2019attentio}, mesh-like connections across multiple layers \cite{cornia2020meshed}, high-order feature interaction \cite{pan2020x}, and relative geometry relationships of objects \cite{guo2020normalized}. Most recently, an auto-parsing network \cite{yang2021auto} is designed to softly segment the inputs into a hierarchical tree, which is further imposed into Transformer-based encoder-decoder for image captioning.

\begin{figure*}[!tb]
\vspace{-0.25in}
    \centering {\includegraphics[width=0.88\textwidth]{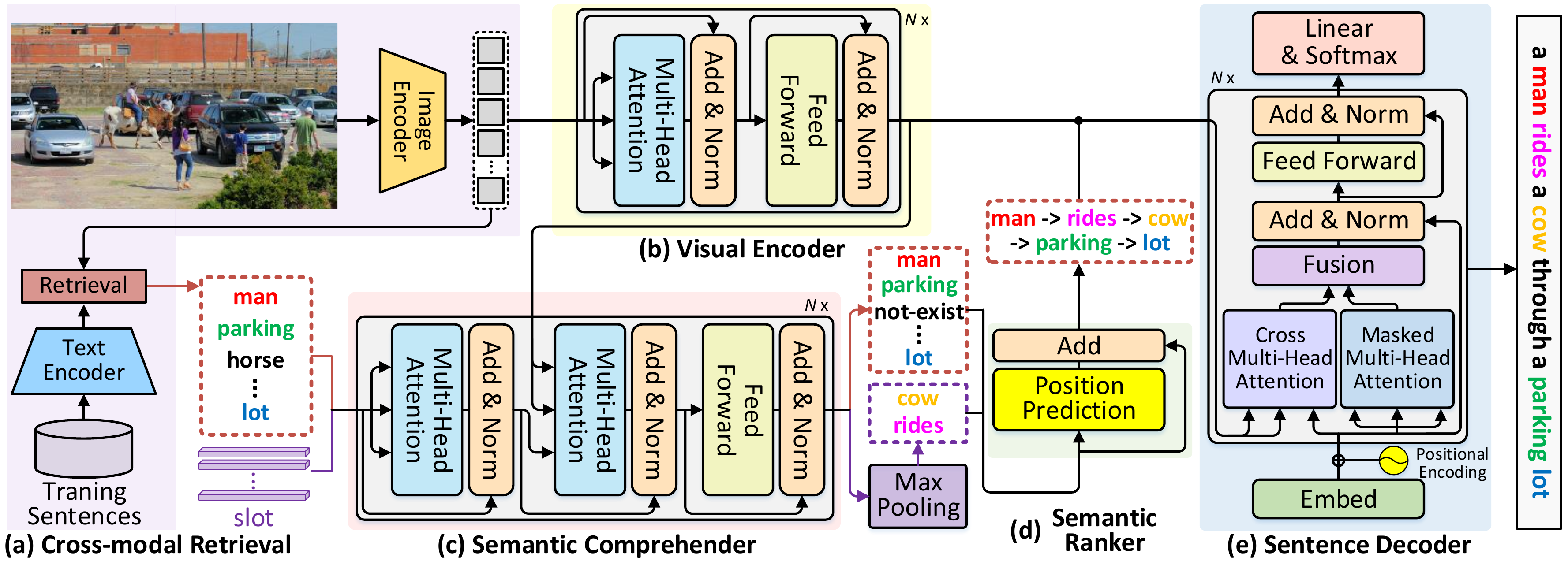}}
    \vspace{-0.12in}
    \caption{An overview of our COS-Net. (a) Given an input image, CLIP first extracts its grid features via image encoder, and then retrieves the semantically similar sentences from sentence pool, which are decomposed into a set of semantic words that act as primary semantic cues. (b) Visual encoder further transforms the grid features into visual tokens through self-attention. (c) Next, semantic comprehender screens the primary semantic cues by filtering out irrelevant semantic words, and meanwhile reconstructs the missing relevant semantic words. (d) The semantic ranker learns to estimate the linguistic position of each semantic word, leading to a sequence of ordered semantic words. (e) Finally, both of visual tokens and ordered semantic words are integrated into sentence decoder for caption generation.}
    \label{fig:framework}
    \vspace{-0.2in}
\end{figure*}

\textbf{Summary.}
The proposed COS-Net can also be considered Transformer-based encoder-decoder scheme that constructs most modules (e.g., visual encoder, sentence decoder, and semantic comprehender) with Transformer-style structure. CLIP-ViL \cite{shen2021much} is perhaps the most related work, which directly takes the pre-trained image encoder in CLIP as visual encoder in Transformer-based encoder-decoder \cite{luo2020better}. Our COS-Net goes beyond CLIP-ViL by utilizing CLIP to seek richer semantic cues that are worthy of mention from human-annotated sentence pool via cross-modal retrieval. Moreover, the semantic comprehender novelly refines the primary semantic cues by filtering out irrelevant semantic words and inferring missing relevant semantic words. A subsequent semantic ranker further allocates all refined semantic words in linguistic order, which serve as additional supervisory signals to boost image captioning.

\section{Our Approach: COS-Net}
Now we proceed to present our core proposal, i.e., Comprehending and Ordering Semantics Networks (COS-Net), that integrates both semantic comprehending and ordering processes into a unified architecture for image captioning. Figure \ref{fig:framework} depicts the detailed architecture of COS-Net.

\subsection{Visual Content Encoding}
Inspired by Transformer-based encoder in image captioning \cite{cornia2020meshed,huang2019attentio} or image recognition \cite{li2021contextual}, we capitalize on multiple stacked Transformer blocks to encode the visual content into intermediate visual tokens. Formally, given an input image $I$, we first employ the image encoder of CLIP \cite{radford2021learning} (backbone: ResNet-101) to extract the grid feature map $\mathcal{V}_I={{\bf{v}}_i}|_{i = 1}^{{N_I}}$ ($N_I$ grids), coupled with the global feature ${\bf{v}}_{c}$. Then, we transform both the global and grid features into a new embedding space, and further concatenate them as: $\mathcal{V}^{(0)}_I = [ {\bf{v}}^{(0)}_{c}, {{\bf{v}}^{(0)}_i}|_{i = 1}^{{N_I}} ]$. After that, a visual encoder is employed to contextually encode all the transformed global and grid features $\mathcal{V}^{(0)}_I$ via self-attention, yielding the enriched visual tokens $\mathcal{V}^{(N_v)}_I = [ {\bf{v}}^{(N_v)}_{c}, {{\bf{v}}^{(N_v)}_i}|_{i = 1}^{{N_I}} ]$. Specifically, we implement this visual encoder by stacking $N_v$ Transformer blocks with multi-head attention. Hence, the $i$-th Transformer block in visual encoder operates as:
\begin{equation}\scriptsize
\label{eq:p1}
\begin{aligned}
&{\mathcal{V}^{(i+1)}_I = \mathcal{F}({\bf{norm}}(\mathcal{V}^{(i)}_I + {\bf{MultiHead}}(\mathcal{V}^{(i)}_I, \mathcal{V}^{(i)}_I, \mathcal{V}^{(i)}_I)))}, \\
&{{\bf{MultiHead}}({\bf{Q}}, {\bf{K}}, {\bf{V}}) = {\bf{Concat}}({head}_{1},...,{head}_{h}){W^O}}, \\
&{{head}_{i} = {\bf{Attention}}({\bf{Q}}W_i^Q, {\bf{K}}W_i^K, {\bf{V}}W_i^V)}, \\
&{ {\bf{Attention}}({\bf{Q}}, {\bf{K}}, {\bf{V}}) = {\bf{softmax}}(\frac{{{\bf{Q}}{{\bf{K}}^T}}}{{\sqrt d }}){\bf{V}}},
\end{aligned}
\end{equation}
where $\mathcal{F}$ denotes the feed-forward layer, $\bf{norm}$ is layer normalization, $W_i^Q$, $W_i^K$, $W_i^V$, $W^O$ are weight matrices, and $d$ is a scaling factor. Note that in order to enable the inter-layer global feature interaction, we additionally concatenate the output global features from all Transformer blocks, which are further transformed into a holistic global feature:
\begin{equation}
\label{eq:p2}
{\bf{\tilde v}}_c = {W_c}[{\bf{v}}_c^{(0)},{\bf{v}}_c^{(1)},...,{\bf{v}}_c^{({N_v})}],
\end{equation}
where $W_c$ is weight matrix. Accordingly, by additionally integrating the encoded grid features of visual encoder with the holistic global feature ${\bf{\tilde v}}_c$, we obtain the final output visual tokens $\mathcal{{\tilde V}}_I=[ {\bf{\tilde v}}_c, {{\bf{v}}^{(N_v)}_i}|_{i = 1}^{{N_I}} ]$.

\subsection{Semantic Comprehending}
Most existing image captioning techniques leverage a pre-trained object detector/classifier to capture the semantics in an image, which are directly fed into sentence decoder to produce the caption. Nevertheless, the semantic perception capability of these pre-trained detector/classifier is severely limited by pre-defined semantic/class labels. Moreover, the separate optimization between pre-trained detector/classifier and sentence decoder hinders the interaction in between. That makes it difficult to adaptively tune the object detector/classifier to better emphasize the salient semantics that are worthy of mention in the output sentence. To alleviate these limitations, we propose to utilize the off-the-shelf CLIP trained on diverse and large-scale data as a powerful cross-modal retrieval model, that directly accumulates more candidates of semantic words that tend to be mentioned in visually similar images. Based on such primary semantic cues mined through cross-modal retrieval, a new semantic comprehender is designed to screen out irrelevant semantic words and meanwhile infer the missing relevant semantic words, pursuing a comprehensive and accurate semantic understanding.

\textbf{Cross-modal Retrieval.}
In an effort to exploit the richer contextual semantics implied in existing human-annotated image-sentence pairs in training set, we capitalize on a cross-modal retrieval model (i.e., CLIP) to search semantically relevant sentences in training sentence pool for each input image. Technically, let ${\bf{v}}_{c}$ and ${\bf{s}}^{c}$ denote the visual and textual feature extracted by the image encoder~and text encoder in CLIP for the input image ${I}$ and each sentence $\mathcal{S}$, respectively. Thus, by taking the input image $I$ as the search query, we retrieve the top-$K$ captions $\mathcal{S}_r = \{\mathcal{S}_{r_1}, \mathcal{S}_{r_2},..., \mathcal{S}_{r_K}\}$ from training sentence pool according to the cosine similarity between $I$ and each caption~$\mathcal{S}_{r_k}$:
\begin{equation}
\label{eq:p3}
{\bf{Similarity}}(I,{\mathcal{S}_{{r_k}}}) = \frac{{{{\bf{v}}_{c}} \cdot {\bf{s}}_{{r_k}}^{c}}}{{||{{\bf{v}}_{c}}||\,||{\bf{s}}_{{r_k}}^{c}||}} ,
\end{equation}
where ${\bf{s}}_{{r_k}}^{c}$ is the textual feature of caption $\mathcal{S}_{r_k}$. After obtaining all the $K$ searched captions that are semantically relevant to the input image, we decompose them into~a set of $N_r$ semantic words $\mathcal{V}_s={{\bf{s}}_i}|_{i = 1}^{{N_r}}$ by removing the stop words, which are further taken as the primary semantic~cues.

\textbf{Semantic Comprehender.}
Although the primary semantic cues derived from cross-modal retrieval cover more relevant semantic words that are worthy of mention, more irrelevant semantic words are also inevitably introduced. A semantic comprehender is thus utilized to filter out the irrelevant semantic words and meanwhile enrich the primary semantic cues with more relevant but missing semantic words. Concretely, we formulate such process of semantic screening and enriching as a set prediction problem \cite{carion2020end}, which directly transforms the primary semantic cues $\mathcal{V}_s={{\bf{s}}_i}|_{i = 1}^{{N_r}}$ into the refined semantic predictions conditioned on the visual tokens $\mathcal{{\tilde V}}_I$. Note that in order to enable the reconstruction of the missing relevant semantic words, we augment the inputs of primary semantic cues $\mathcal{V}_s$ with the additional parametric semantic queries (i.e., a set of slots $\mathcal{O}={{\bf{o}}^{(0)}_i}|_{i = 1}^{{N_o}}$). More specifically, the primary semantic cues $\mathcal{V}_s$ is first mapped into a new semantic embedding space, leading to the primary semantic features ${{\bf{s}}^{(0)}_i}|_{i = 1}^{{N_r}}$. Next, we feed the concatenation of primary semantic features and the parametric semantic queries (i.e., $\mathcal{V}^{(0)}_s=[{{\bf{o}}^{(0)}_i}|_{i = 1}^{{N_o}}, {{\bf{s}}^{(0)}_i}|_{i = 1}^{{N_r}}]$) into semantic comprehender to trigger the set prediction of the screened and enriched semantic words. Here we implement the semantic comprehender as $N_s$ stacked Transformer blocks. Each block contextually encodes every input semantic word (i.e., semantic token) via self-attention, and further enhances the semantic tokens by exploiting the interaction between them and visual tokens $\mathcal{{\tilde V}}_I$ via cross-attention, which is measured as:
\begin{equation}\small
\label{eq:p4}
\begin{aligned}
{\mathcal{V}^{(i+1)}_s = \mathcal{F}({\bf{norm}}(\mathcal{V}'_s + {\bf{MultiHead}}(\mathcal{V}'_s, \mathcal{{\tilde V}}_I, \mathcal{{\tilde V}}_I)))}, \\
{\mathcal{V}'_s = {\bf{norm}}(\mathcal{V}^{(i)}_s + {\bf{MultiHead}}(\mathcal{V}^{(i)}_s, \mathcal{V}^{(i)}_s, \mathcal{V}^{(i)}_s))}, \\
\end{aligned}
\end{equation}
where $\mathcal{V}^{(i+1)}_s$ denotes the output enhanced semantic tokens of $i$-th Transformer block. Accordingly, the final output semantic tokens of semantic comprehender $\mathcal{V}^{(N_s)}_s=[{{\bf{o}}^{(N_s)}_i}|_{i = 1}^{{N_o}}, {{\bf{s}}^{(N_s)}_i}|_{i = 1}^{{N_r}}]$, are leveraged for predicting the refined and reconstructed semantic words.

\textbf{Objective.}
During training, we include a proxy objective to optimize the semantic comprehender by encouraging the filter of irrelevant semantic words in primary semantic cues and the reconstruction of the missing relevant semantic words. Here we formulate this process as a combination of single-label and multi-label classification problems. In particular, conditioned on the output semantic tokens of semantic comprehender $\mathcal{V}^{(N_s)}_s=[{{\bf{o}}^{(N_s)}_i}|_{i = 1}^{{N_o}}, {{\bf{s}}^{(N_s)}_i}|_{i = 1}^{{N_r}}]$, a prediction layer is utilized to estimate the probability distribution over the whole semantic vocabulary for each semantic token, yielding the semantic predictions $\mathcal{P}_s = [{P_{o_i}}|_{i = 1}^{{N_o}}, {P_{s_i}}|_{i = 1}^{{N_r}}]$. Note that the semantic vocabulary is constructed as all the $N_c$ semantic words in training set plus one special token that represents irrelevant semantic word. The ground-truth label for the prediction of $i$-th semantic token ${P_{s_i}}$ in primary semantic cues is thus denoted as $y_i \in {\mathbb{R}}^{N_c+1}$. In this way, based on ${P_{s_i}}|_{i = 1}^{{N_r}}$, we treat the process of filtering out irrelevant semantic words in primary semantic cues as the task of single-label classification, and its objective is measured with cross-entropy loss:
\begin{equation}
\label{eq:p6_1}
\begin{aligned}
{\mathcal{L}_x} = - \frac{1}{{_{{N_r}}}}\sum\limits_{i = 1}^{{N_r}} {\sum\limits_{c = 1}^{{N_c} + 1} {y_i^c\log P_{{s_i}}^c} },
\end{aligned}
\end{equation}
where $y_i^c$ and $P_{s_i}^c$ denotes the $c$-th element of $y_i$ and $P_{s_i}$, respectively.
Meanwhile, we regard the process of inferring the missing relevant semantic words as the task of multi-label classification. Specifically, after normalizing the predictions of parametric semantic queries ${{P_{o_i}}}|_{i = 1}^{{N_o}}$ with $\bf{sigmoid}$ activation, we perform max pooling over them to achieve the holistic probability distribution $\tilde P_o$ over semantic vocabulary. Therefore, the objective of multi-label classification is calculated with asymmetric loss \cite{ridnik2021asymmetric}:
\begin{equation}
\label{eq:p6_2}
\begin{aligned}
{\mathcal{L}_m} = {\bf{asym}}({\tilde P_o}, {\bf{y}}_m),
\end{aligned}
\end{equation}
where $\bf{asym}$ denotes the asymmetric loss and ${\bf{y}}_m$ is the ground-truth label of all missing relevant semantic words. Finally, the whole objective of semantic comprehender integrates both objectives of filtering out irrelevant semantic words and reconstructing missing relevant semantic words:
\begin{equation}
\label{eq:p6_3}
\begin{aligned}
{\mathcal{L}_s} = \mathcal{L}_x + \mathcal{L}_m.
\end{aligned}
\end{equation}

\subsection{Semantic Ordering}
After obtaining the screened and enriched semantics derived from semantic comprehender, the most typical way to generate description is to directly feed them into RNN/Transformer based sentence decoder for sentence modeling. However, this way overly relies on the language priors, possibly resulting in non-existent semantic words due to the phenomenon of object hallucination. To address the issue, we additionally involve a new module of semantic ranker that learns to estimate the linguistic position of each semantic word, thereby allocating all the semantic words in linguistic order as humans. In this way, the output sequence of ordered semantic words serve as additional visually-grounded language priors to encourage the generation of both relevant and coherent descriptions.

Conventional Transformer encoder-decoder characters the linguistic order of each word through a static learnable encoding of pre-defined position in a sequence. Nevertheless, in our context, the specific position of each semantic word is unclear after semantic comprehending, and the inherent correspondence between each semantic word and its linguistic order is dynamic. Therefore, instead of representing each linguistic order as a static position encoding, our semantic ranker capitalizes on attention mechanism to dynamically infer the linguistic position of each semantic word. Formally, we first initialize a set of $D$-dimensional position encodings $\mathcal{V}_p \in {\mathbb{R}}^{{N_p} \times {D}} $ that depict all linguistic orders in a sequence, where $N_p$ is the maximum length of semantic word sequence. Next, for each semantic word (e.g., the $i$-th semantic token ${\tilde v}_{s_i}$ in $\mathcal{V}^{(N_s)}_s$), we measure its attention distribution over all position encodings $\mathcal{V}_p$ and then calculate its attended position encoding by aggregating all position encodings with attention:
\begin{equation}
\label{eq:p5}
{p_i = {\bf{softmax}}({\tilde v}_{s_i} {\mathcal{V}_p}^T){\mathcal{V}_p}}.
\end{equation}
Here the attended position encoding $p_i$ can be interpreted as a ``soft" estimation of the linguistic order of each semantic token ${\tilde v}_{s_i}$ in the semantic word sequence. After that, we upgrade each semantic token with its estimated linguistic order, leading to the position-aware semantic token:
\begin{equation}
\label{eq:p55}
{{\tilde v}_{s_i}^p = {\tilde v}_{s_i} + p_i}.
\end{equation}
Accordingly, the semantic ranker produces a set of position-aware semantic tokens $\mathcal{{\tilde V}}_s = \{ {{\tilde v}_{s_1}^p, {\tilde v}_{s_2}^p, ..., {\tilde v}_{s_{N_o+N_r}}^p} \} $ that present the sequence of ordered semantic words.

\subsection{Sentence Decoding}
With the enriched visual tokens $\mathcal{{\tilde V}}_I$ from visual encoder and the position-aware semantic tokens $\mathcal{{\tilde V}}_s$ from semantic ranker, we then discuss how to integrate them into the Transformer-based decoder for sentence generation. Formally, let $\mathcal{S}=\{ {w_0},{w_1},...,{w_{T-1}}\}$ denote the textual sentence ($T$: word number) that describes input image $I$. Each word is represented as a ``one-hot" vector, which is further transferred into a $D$-dimensional textual feature via weight matrix: $H^{(0)}_{0:T-1}=\{{h^{(0)}_0}, {h^{(0)}_1},...,{h^{(0)}_{T-1}} \}$. In general, the sentence decoder takes each word as input and learns to predict the next word auto-regressively conditioned on the enriched visual tokens $\mathcal{{\tilde V}}_I$ and the position-aware semantic tokens $\mathcal{{\tilde V}}_s$. We implement the sentence decoder as $N_d$ stacked Transformer blocks. Each Transformer block is composed of a masked multi-head attention layer to model the holistic textual context of previous generated words, and a cross multi-head attention layer that integrates both visual and semantic tokens to trigger sentence generation. Specifically, at the $t$-th decoding timestep, the masked multi-head attention layer in $i$-th block performs self-attention over previous generated words based on the query of previous output hidden state $h^{(i)}_t$, leading to the holistic textual context $h'^{(i)}_t$:
\begin{equation}\scriptsize
\label{eq:p7}
h'^{(i)}_t = {\bf{MultiHead}}(h^{(i)}_t, H^{(i)}_{0:t}, H^{(i)}_{0:t}).
\end{equation}
After that, the cross multi-head attention layer is employed to separately conduct cross-attention over the visual tokens $\mathcal{{\tilde V}}_I$ and the semantic tokens $\mathcal{{\tilde V}}_s$ depending on the same query (i.e., $h^{(i)}_t$), yielding the holistic visual context $h^{v{(i)}}_t$:
\begin{equation} \scriptsize
\label{eq:p8}
h^{v{(i)}}_t = {\bf{MultiHead}}(h^{(i)}_t, \mathcal{{\tilde V}}_I, \mathcal{{\tilde V}}_I) + {\bf{MultiHead}}(h^{(i)}_t, \mathcal{{\tilde V}}_s, \mathcal{{\tilde V}}_s).
\end{equation}
Next, we fuse the holistic textual context $h'^{(i)}_t$ and visual context $h^{v{(i)}}_t$ with a $\bf{sigmoid}$ gate function, and the learnt hidden state $h^{(i+1)}_t$ is taken as the outputs of $i$-th block:
\begin{equation} \scriptsize
\label{eq:p9}
\begin{aligned}
&{h^{(i+1)}_t = \mathcal{F}({\bf{norm}}(h^{(i)}_t + (g * h'^{(i)}_t + (1-g) * h^{v{(i)}}_t)))},\\
&{g = {\bf{Sigmoid}}(W_g [h^{v{(i)}}_t, h'^{(i)}_t])}.
\end{aligned}
\end{equation}
Finally, the output hidden state of the last block $h^{(N_d)}_t$ is utilized for predicting the next word ${w_{t+1}}$ via softmax.

\subsection{Overall Objective}
At training stage, the overall objective of our COS-Net is measured as the integration of the proxy objective in semantic comprehender $L_s$ and the typical cross entropy loss $L_{XE}$ for sentence generation:
$\mathcal{L}=\mathcal{L}_{s}+\mathcal{L}_{XE}$.
Next, following \cite{luo2020better}, COS-Net can be further optimized with sentence-level reward (e.g., CIDEr score).

\section{Experiments}

\subsection{Dataset and Experimental Settings}

\textbf{Dataset.}
We empirically verify and analyze the effectiveness of our COS-Net on the widely adopted COCO benchmark \cite{Lin:ECCV14} for image captioning. The COCO dataset consists of more than 120,000 images, and each image is equipped with five human-annotated sentences. For fair comparison with most existing techniques, we strictly follow the standard dataset split in \cite{Karpathy:CVPR15} (known as Karpathy split), which leverages 5,000 images for validation, 5,000 images for testing, and the rest for training. Besides the standard Karpathy split, we adopt the robust split introduced in \cite{lu2018neural} to conduct object hallucination analysis, which ensures that the object pairs mentioned in training, validation, and testing captions do not overlap.
In the experiments, we perform the minimal sentence pre-processing by converting each sentence into lower case and meanwhile filtering out rare words that occur less than six times as in \cite{anderson2017bottom}. The overall word vocabulary is thus built with 10,199 unique words. Moreover, to enable the learning of our semantic comprehender, we construct an additional semantic vocabulary ($N_c=906$) by removing all the stop words in word vocabulary and selecting high-frequency semantic words.

\begin{table}[t]\small
  \centering
  \vspace{-0.00in}
  \setlength\tabcolsep{1.8pt}
  \caption{\small Ablation study for COS-Net on COCO Karpathy test split. \textbf{Base}: A base Transformer-based encoder-decoder structure by using CLIP grid features as visual inputs; \textbf{CR}: Cross-modal Retrieval; \textbf{FIS}: Filtering out Irrelevant Semantics; \textbf{IMS}: Inferring Missing Semantics; \textbf{SR}: Semantic Ranker.}
  \vspace{-0.1in}
  \begin{tabular}{c|ccccc|ccccccc}
  \Xhline{2\arrayrulewidth}
  \# & Base & CR          & FIS          & IMS          & SR           & B@4  & M    & R    & C     & S & CHs  & CHi   \\ \hline\hline
  1  & $\checkmark$ &              &              &              &              & 38.0 & 29.0 & 57.9 & 123.6 & 22.1 & 6.2 & 4.3 \\
  2  & $\checkmark$ & $\checkmark$ &              &              &              & 38.4 & 29.3 & 58.5 & 124.9 & 22.3 & 5.3 & 3.6 \\
  3  & $\checkmark$ & $\checkmark$ & $\checkmark$ &              &              & 38.6 & 29.3 & 58.5 & 125.8 & 22.4 & 5.2 & 3.6 \\
  4  & $\checkmark$ & $\checkmark$ & $\checkmark$ & $\checkmark$ &              & \textbf{39.2} & 29.5 & 58.7 & 126.1 & 22.6 & 5.1 & 3.5 \\
  5  & $\checkmark$ & $\checkmark$ & $\checkmark$ & $\checkmark$ & $\checkmark$ & \textbf{39.2} & \textbf{29.7} & \textbf{58.9} & \textbf{127.4} & \textbf{22.7} & \textbf{4.7} & \textbf{3.2} \\
  \Xhline{2\arrayrulewidth}
  \end{tabular}
  \vspace{-0.2in}
  \label{tab:COCO_ablation}
\end{table}

\textbf{Implementation Details.}
In COS-Net, the visual encoder, semantic comprehender, and sentence decoder are constructed with $N_v=6$, $N_s=3$, and $N_d=6$ Transformer blocks (hidden state size: 512). The image encoder in CLIP \cite{radford2021learning} is directly employed over the input image, and each image is thus represented as a 512-dimensional global feature vector plus the 2,048-dimensional grid feature map. The typical two-stage training paradigm \cite{rennie2017self} is adopted to train COS-Net. The whole architecture is implemented based on X-modaler codebase \cite{li2021x}. Specifically, we first optimize the whole architecture of COS-Net by integrating the cross entropy loss with the proxy objective of semantic comprehender for 30 epoches (batch size: 32). In this stage, we leverage Adam \cite{kingma2014adam} optimizer with the learning rate scheduling strategy in \cite{vaswani2017attention} (warmup: 20,000 iterations). For the second stage, we further optimize COS-Net with CIDEr score via self-critical sequence training strategy \cite{luo2020better} for another 50 epoches. The learning rate is set as 0.00001. At inference, the beam size in beam search strategy is set as 3. Following the standard evaluation setup, we report the performances of COS-Net over five evaluation metrics: BLEU@N \cite{Papineni:ACL02} (B@1-4), METEOR \cite{Banerjee:ACL05} (M),~ROUGE \cite{lin2004rouge} (R), CIDEr \cite{vedantam2015cider} (C), and SPICE \cite{spice2016} (S). In addition, we use CHAIR metric \cite{rohrbach2018object} to assess the rate of object hallucination on the robust split. CHAIR metric includes two variants: CHAIRi (CHi) that measures what fraction of objects are hallucinated, and CHAIRs (CHs) that calculates what fraction of sentences include a hallucinated object.

\begin{table*}[t]\scriptsize
  \centering
  \vspace{-0.20in}
  \setlength\tabcolsep{4.1pt}
  \caption{\small The performances of various methods on COCO Karpathy test split (single model setup). \textdagger~denotes our implementations by using CLIP grid features (backbone: ResNet-101) as visual inputs. \textasteriskcentered~utilizes CLIP grid features in a superior backbone (ResNet-50$\times$4).}
  \vspace{-0.1in}
  \begin{tabular}{l|cccccccc|cccccccc}
  \Xhline{2\arrayrulewidth}
               & \multicolumn{8}{c|}{\textbf{Cross-Entropy Loss}} & \multicolumn{8}{c}{\textbf{CIDEr Score Optimization}} \\
               & B@1  & B@2  & B@3 & B@4 & M & R & C & S & B@1  & B@2  & B@3  & B@4  & M  & R  & C  & S \\ \hline\hline
Up-Down \cite{anderson2017bottom} & 77.2 &   -  &   -  & 36.2 & 27.0 & 56.4 & 113.5 & 20.3 & 79.8 &  -   &  -   & 36.3 & 27.7 & 56.9 & 120.1 & 21.4 \\
GCN-LSTM \cite{yao2018exploring}  & 77.3 &   -  &   -  & 36.8 & 27.9 & 57.0 & 116.3 & 20.9 & 80.5 &  -   &  -   & 38.2 & 28.5 & 58.3 & 127.6 & 22.0 \\
SGAE \cite{Yang:CVPR19}  & 77.6 &   -  &   -  & 36.9 & 27.7 & 57.2 & 116.7 & 20.9 & 80.8 &  -   &  -   & 38.4 & 28.4 & 58.6 & 127.8 & 22.1 \\
AoANet \cite{huang2019attentio}   & 77.4 &   -  &   -  & 37.2 & 28.4 & 57.5 & 119.8 & 21.3 & 80.2 &  -   &  -   & 38.9 & 29.2 & 58.8 & 129.8 & 22.4 \\
Transformer \cite{sharma2018conceptual} & 76.4 & 60.3 & 46.5 & 35.8 & 28.2 & 56.7 & 116.6 & 21.3 & 80.5 & 65.4 & 51.1 & 39.2 & 29.1 & 58.7 & 130.0 & 23.0 \\
$M^2$ Transformer \cite{cornia2020meshed} &  - & -  & -  & -  &   -  &  -   &   -   &   -  & 80.8 &  -   &  -   & 39.1 & 29.2 & 58.6 & 131.2 & 22.6 \\
APN \cite{yang2021auto}         &  - & -  & -  & -  &   -  &  -   &   -   &   -  & - & - & - & 39.6 & 29.2 & 59.1 & 131.8 & 23.0 \\
NG-SAN \cite{guo2020normalized} &  - & -  & -  & -  &   -  &  -   &   -   &   -  & - & - & - & 39.9 & 29.3 & 59.2 & 132.1 & 23.3 \\
X-Transformer \cite{pan2020x} & 77.3 & 61.5 & 47.8 & 37.0 & 28.7 & 57.5 & 120.0 & 21.8 & 80.9 & 65.8 & 51.5 & 39.7 & 29.5 & 59.1 & 132.8 & 23.4 \\ \hline
CLIP-Res101  \cite{shen2021much} &  - & -  & -  & -  &   -  &  -   &   -   &   -  & - & - & - & 39.2 & 29.1 & - & 130.3 & 23.0 \\
CLIP-Res50$\times$4 \textasteriskcentered \cite{shen2021much} &  - & -  & -  & -  &   -  &  -   &   -   &   -  & - & - & - & 40.2 & 29.7 & - & 134.2 & 23.8 \\
Up-Down \textdagger \cite{anderson2017bottom} & 78.1 & 62.6 & 49.1 & 38.3 & 28.6 & 57.9 & 120.7 & 21.6 & 81.3 & 66.2 & 51.5 & 39.4 & 29.2 & 59.3 & 131.9 & 22.8 \\
Transformer \textdagger \cite{sharma2018conceptual} & 78.0 & 62.4 & 48.9 & 38.0 & 29.0 & 57.9 & 123.6 & 22.1 & 81.6 & 66.9 & 52.6 & 40.6 & 29.9 & 59.8 & 136.2 & 23.9 \\
X-Transformer \textdagger \cite{pan2020x} & 78.3 & 62.9 & 49.3 & 38.2 & 29.2 & 58.3 & 124.5 & 22.6 & 82.0 & 67.2 & 53.1 & 41.2 & 30.2 & 60.0 & 137.2 & 24.2 \\ \hline
COS-Net                       & \textbf{79.2} & \textbf{63.8} & \textbf{50.2} & \textbf{39.2} & \textbf{29.7} & \textbf{58.9} & \textbf{127.4} & \textbf{22.7} & \textbf{82.7} & \textbf{68.2} & \textbf{54.0} & \textbf{42.0} & \textbf{30.6} & \textbf{60.6} & \textbf{141.1} & \textbf{24.6} \\
  \Xhline{2\arrayrulewidth}
  \end{tabular}
  \vspace{-0.1in}
  \label{tab:COCO_single}
\end{table*}

\begin{table*}[t]\scriptsize
  \centering
  \setlength\tabcolsep{4.1pt}
  \caption{\small The performances of various methods on COCO Karpathy test split (ensemble model setup).}
  \vspace{-0.1in}
  \begin{tabular}{l|cccccccc|cccccccc}
  \Xhline{2\arrayrulewidth}
               & \multicolumn{8}{c|}{\textbf{Cross-Entropy Loss}} & \multicolumn{8}{c}{\textbf{CIDEr Score Optimization}} \\
               & B@1  & B@2  & B@3 & B@4 & M & R & C & S & B@1  & B@2  & B@3  & B@4  & M  & R  & C  & S \\ \hline\hline
GCN-LSTM \cite{yao2018exploring}  & 77.4 & - & - & 37.1 & 28.1 & 57.2 & 117.1 & 21.1 & 80.9 &  -   &  -   & 38.3 & 28.6 & 58.5 & 128.7 & 22.1 \\
SGAE \cite{Yang:CVPR19}  &  -   &   -  &   -  &  -   &   -  &   -  &   -   &   -  & 81.0 &  -   &  -   & 39.0 & 28.4 & 58.9 & 129.1 & 22.2 \\
AoANet \cite{huang2019attentio}   & 78.7 & - & - & 38.1 & 28.5 & 58.2 & 122.7 & 21.7 & 81.6 &  -   &  -   & 40.2 & 29.3 & 59.4 & 132.0 & 22.8 \\
$M^2$ Transformer \cite{cornia2020meshed} &  - & -  & -  & -  &   -  &  -   &   -   &   -  & 82.0 &  -   &  -   & 40.5 & 29.7 & 59.5 & 134.5 & 23.5 \\
X-Transformer \cite{pan2020x} & 77.8 & 62.1 &  48.6 & 37.7 & 29.0 & 58.0 & 122.1 & 21.9 & 81.7 & 66.8 & 52.6 & 40.7 & 29.9 & 59.7 & 135.3 & 23.8 \\ \hline
COS-Net        & \textbf{79.6} &  \textbf{64.4} &  \textbf{50.9} &  \textbf{40.0} & \textbf{30.0}  &  \textbf{59.4} & \textbf{129.5} & \textbf{22.9} & \textbf{83.5} & \textbf{69.1} & \textbf{54.9} & \textbf{42.9} & \textbf{30.8} & \textbf{61.0} & \textbf{143.0} & \textbf{24.7}  \\

  \Xhline{2\arrayrulewidth}
  \end{tabular}
  \vspace{-0.2in}
  \label{tab:COCO_ensemble}
\end{table*}

\subsection{Ablation Study}
In this section, we conduct ablation study to investigate how each design in our COS-Net influences the overall performances on COCO dataset. Table \ref{tab:COCO_ablation} details the performance comparisons among different ablated runs of our COS-Net. Note that all results here are reported without self-critical sequence training strategy. We start from a base Transformer-based encoder-decoder structure (\textbf{Base}), which is a degraded version of COS-Net by solely using the CLIP grid features as visual inputs, without exploring primary semantic cues via cross-modal retrieval, semantic comprehending and ordering. After that, we extend the Based model by additionally exploring CLIP as cross-modal retrieval model to mine the primary semantic cues for boosting sentence generation. In this way, \textbf{Base+CR} exhibits better performances, which verify the merit of accumulating richer semantic words that tend to be mentioned in visually similar images through cross-modal retrieval. Next, \textbf{Base+CR+FIS} learns to filter out the irrelevant semantic words in primary semantic cues, and thus leads to performance gains. \textbf{Base+CR+FIS+IMS} is further benefited from the additional process of inferring the missing relevant semantic words. The results of these two ablated runs basically highlight the advantage of semantic screening and enriching in our semantic comprehender for image captioning. Finally, after integrating Base+CR+FIS+IMS with our semantic ranker that estimates the linguistic position of each semantic word derived from semantic comprehender, \textbf{Base+CR+FIS+IMS+SR} (i.e., our COS-Net) achieves the best performances across most evaluation metrics. The results validate the leverage of the sequence of ordered semantic words as additional visually-grounded language priors to enhance sentence generation.

\subsection{Comparisons with State-of-the-Art}

Here we compare our COS-Net with a series of state-of-the-art image captioning approaches on three different splits, i.e., the standard Karpathy test split, the official test split via online evaluation, and the robust split for object hallucination analysis. Specifically, for Karpathy test split, we follow modern techniques and utilize two different training setups for evaluation. One is the default single model setup that produces sentence via a single model, and the other is ensemble model setup that ensembles multiple models with different initialized parameters.

\textbf{Single Model on Karpathy Test Split.}
Table \ref{tab:COCO_single} summarizes the performance comparisons in the default single model setup. All runs are briefly grouped into two directions: (1) the standard methods (e.g., SGAE \cite{Yang:CVPR19}, Up-Down \cite{anderson2017bottom}, Transformer \cite{sharma2018conceptual}, $M^2$ Transformer \cite{cornia2020meshed}) that utilizes the pre-trained Faster R-CNN (backbone: ResNet-101) to extract visual inputs; (2) the approaches (e.g., CLIP-Res101 \cite{shen2021much}) that take the strong CLIP grid features as visual inputs. Note that for fair comparisons with our COS-Net, we re-implement several upgraded variants of existing standard methods (e.g., Up-Down \textdagger, Transformer~\textdagger, X-Transformer~\textdagger) by using the same CLIP grid features as visual inputs. As shown in this table, our COS-Net consistently outperforms the state-of-the-art methods across all the evaluation metrics. In particular, under the setting of CIDEr score optimization, the CIDEr Score of COS-Net can reach 141.1\%, which leads to the absolute improvement of 3.9\% against the best competitor X-Transformer~\textdagger~(CIDEr: 137.2\%). This generally demonstrates the key advantage of jointly comprehending and ordering the semantics in an image to facilitate sentence generation. Compared to the methods that leverage RNN-based structure (e.g., Up-Down and GCN-LSTM), Transformer and $M^2$ Transformer improve the performances by utilizing Transformer-based scheme that strengthens vision-language interaction via cross-attention. Instead of using the pre-trained Faster R-CNN to encode visual content in primary Up-Down, Up-Down \textdagger~utilizes the CLIP grid features to trigger bottom-up and top-down attention, leading to clear performance boosts. The results indicate the stronger capability of semantic comprehending in CLIP that is trained on diverse and large-scale data. When further upgrading the conventional Transformer with CLIP grid features, Transformer \textdagger~also manages to achieve better performances. However, these upgraded runs of existing approaches solely hinge on the visual content encoding via pre-trained CLIP without any interaction between CLIP and sentence decoder, and meanwhile ignore the inherent linguistic ordering of semantics. As an alternative, our COS-Net encourages a more comprehensive and accurate semantic understanding, and further learns to allocate the semantic words in linguistic ordering as humans, thereby achieving the best performances in terms of all evaluation metrics.

\textbf{Ensemble Model on Karpathy Test Split.}
Next, we evaluate our COS-Net with ensembles of four models, which are trained with different random seeds. As shown in Table \ref{tab:COCO_ensemble}, the performance trends in the ensemble model setup are similar to those in single model setup. Concretely, the ensemble version of COS-Net surpasses the current state-of-the-art standard technique (ensemble X-Transformer) by an absolute improvement of 7.7\% in CIDEr score. The results again demonstrate the effectiveness of jointly screening \& enriching the primary semantic cues and further ordering semantics for image captioning.

\begin{table*}[!tb]\scriptsize
  \centering
  \vspace{-0.10in}
  \caption{\small The performances of various methods on the official test split in online test server.}
  \label{table:leaderboard}
  \vspace{-0.10in}
  \begin{tabular}{l|*{13}{c|}c}
  \Xhline{2\arrayrulewidth}
      \multicolumn{1}{c|}{\multirow{2}{*}{{Model}}} & \multicolumn{2}{c|}{{B@1}} & \multicolumn{2}{c|}{{B@2}} & \multicolumn{2}{c|}{{B@3}} & \multicolumn{2}{c|}{{B@4}} & \multicolumn{2}{c|}{{M}} & \multicolumn{2}{c|}{{R}} & \multicolumn{2}{c}{{C}} \\\cline{2-15}
      \multicolumn{1}{c|}{}&c5 &c40 &c5 &c40 &c5 &c40&c5 &c40&c5 &c40&c5 &c40&c5 &c40 \\\hline
  	  {Up-Down} \cite{anderson2017bottom}     & 80.2 & 95.2 & 64.1 & 88.8 & 49.1 & 79.4 & 36.9 & 68.5 & 27.6 & 36.7 & 57.1 & 72.4 & 117.9 & 120.5  \\
  {SGAE} \cite{Yang:CVPR19}                   & 81.0 & 95.3 & 65.6 & 89.5 & 50.7 & 80.4 & 38.5 & 69.7 & 28.2 & 37.2 & 58.6 & 73.6 & 123.8 & 126.5 \\
  {GCN-LSTM} \cite{yao2018exploring}          & 80.8 & 95.2 & 65.5 & 89.3 & 50.8 & 80.3 & 38.7 & 69.7 & 28.5 & 37.6 & 58.5 & 73.4 & 125.3 & 126.5 \\
  APN \cite{yang2021auto}                     &  -   &  -   &  -   &  -   &  -   &  -   & 38.9 & 70.2 & 28.8 & 38.0 & 58.7 & 73.7 & 126.3 & 127.6 \\
  {AoANet} \cite{huang2019attentio}           & 81.0 & 95.0 & 65.8 & 89.6 & 51.4 & 81.3 & 39.4 & 71.2 & 29.1 & 38.5 & 58.9 & 74.5 & 126.9 & 129.6 \\
  {X-Transformer} \cite{pan2020x}             & 81.3 & 95.4 & 66.3 & 90.0 & 51.9 & 81.7 & 39.9 & 71.8 & 29.5 & 39.0 & 59.3 & 74.9 & 129.3 & 131.4 \\
  {$M^2$ Transformer} \cite{cornia2020meshed} & 81.6 & 96.0 & 66.4 & 90.8 & 51.8 & 82.7 & 39.7 & 72.8 & 29.4 & 39.0 & 59.2 & 74.8 & 129.3 & 132.1 \\ \hline
  COS-Net                                     & \textbf{83.3} & \textbf{96.8} & \textbf{68.6} & \textbf{92.3} & \textbf{54.2} & \textbf{84.5} & \textbf{42.0} & \textbf{74.7} & \textbf{30.4} & \textbf{40.1} & \textbf{60.6} & \textbf{76.4} & \textbf{136.7} & \textbf{138.3} \\

  \Xhline{2\arrayrulewidth}
  \end{tabular}
  \vspace{-0.18in}
\end{table*}

\begin{table}[t]\small
  \centering
  \setlength\tabcolsep{2.1pt}
  \caption{\small Hallucination analysis on the robust split. \textdagger~denotes our implementations by using CLIP grid features as visual inputs.}
  \vspace{-0.1in}
  \begin{tabular}{l|cccccccc}
  \Xhline{2\arrayrulewidth}
               & B@1 & B@4 & M & R & C & S & CHs  & CHi \\ \hline\hline
Att2In  \cite{rennie2017self}                             &  -   &  -   & 24.0 &  -   & 85.8  & 16.9 & 14.1 & 10.1  \\
Up-Down  \cite{anderson2017bottom}                        &  -   &  -   & 24.7 &  -   & 89.8  & 17.7 & 11.3 & 7.9   \\ \hline
Att2In \textdagger \cite{rennie2017self}                  & 76.5 & 35.7 & 26.7 & 55.7 & 104.4 & 19.8 & 9.0  & 5.9   \\
Up-Down \textdagger \cite{anderson2017bottom}             & 76.8 & 36.3 & 27.1 & 56.0 & 106.3 & 20.1 & 8.6  & 5.6   \\
Transformer \textdagger \cite{sharma2018conceptual}       & 76.9 & 36.3 & 27.4 & 56.1 & 109.3 & 20.5 & 7.9  & 5.1   \\ \hline
COS-Net                                                   & \textbf{78.0} & \textbf{37.3} & \textbf{27.9} & \textbf{56.8} & \textbf{112.1} & \textbf{21.2} & \textbf{6.2}  & \textbf{3.9}   \\
  \Xhline{2\arrayrulewidth}
  \end{tabular}
  \vspace{-0.2in}
  \label{tab:COCO_hallucination}
\end{table}

\textbf{Online Evaluation on Official Test Split.}
We further include more evaluations on the official test split by submitting COS-Net to online test server. Table \ref{table:leaderboard} shows the performances with regard to 5 reference captions (c5) and 40 reference captions (c40). Since most top-performing methods in this online leaderboard adopt the ensemble model setup, here we report the performances of the ensemble COS-Net for fair comparison. Similarly, COS-Net surpasses all state-of-the-art approaches across all metrics.

\textbf{Hallucination Analysis on Robust Split.}
To better understand the impact of semantic comprehending and ordering in our COS-Net, we conduct hallucination analysis \cite{rohrbach2018object} to assess the rate of object hallucination (i.e., the image relevance of the generated captions) on the robust split. Table \ref{tab:COCO_hallucination} lists the performances over both typical sentence metrics and the image relevance metrics (CHs and CHi). Following the evaluation in single model setup, we include two groups of baselines (i.e., the standard methods and their upgraded version with CLIP grid features). Similar trends are also observed in this hallucination analysis. Specifically, by equipping the standard approaches (e.g., Att2In and Up-Down) with CLIP grid features, Att2In~\textdagger~and Up-Down~\textdagger~achieve lower CHs
and CHi scores, which show the stronger semantic understanding capability of CLIP. Moreover, our COS-Net goes beyond Transformer \textdagger~by additionally mining primary semantic cues via cross-modal retrieval and further refining \& ordering the semantics, leading to lower CHs and CHi scores. The results confirm that COS-Net is more robust by alleviating object hallucination.

\subsection{Qualitative Results}

\begin{figure}[!tb]
\vspace{-0.0in}
\centering {\includegraphics[width=0.43\textwidth]{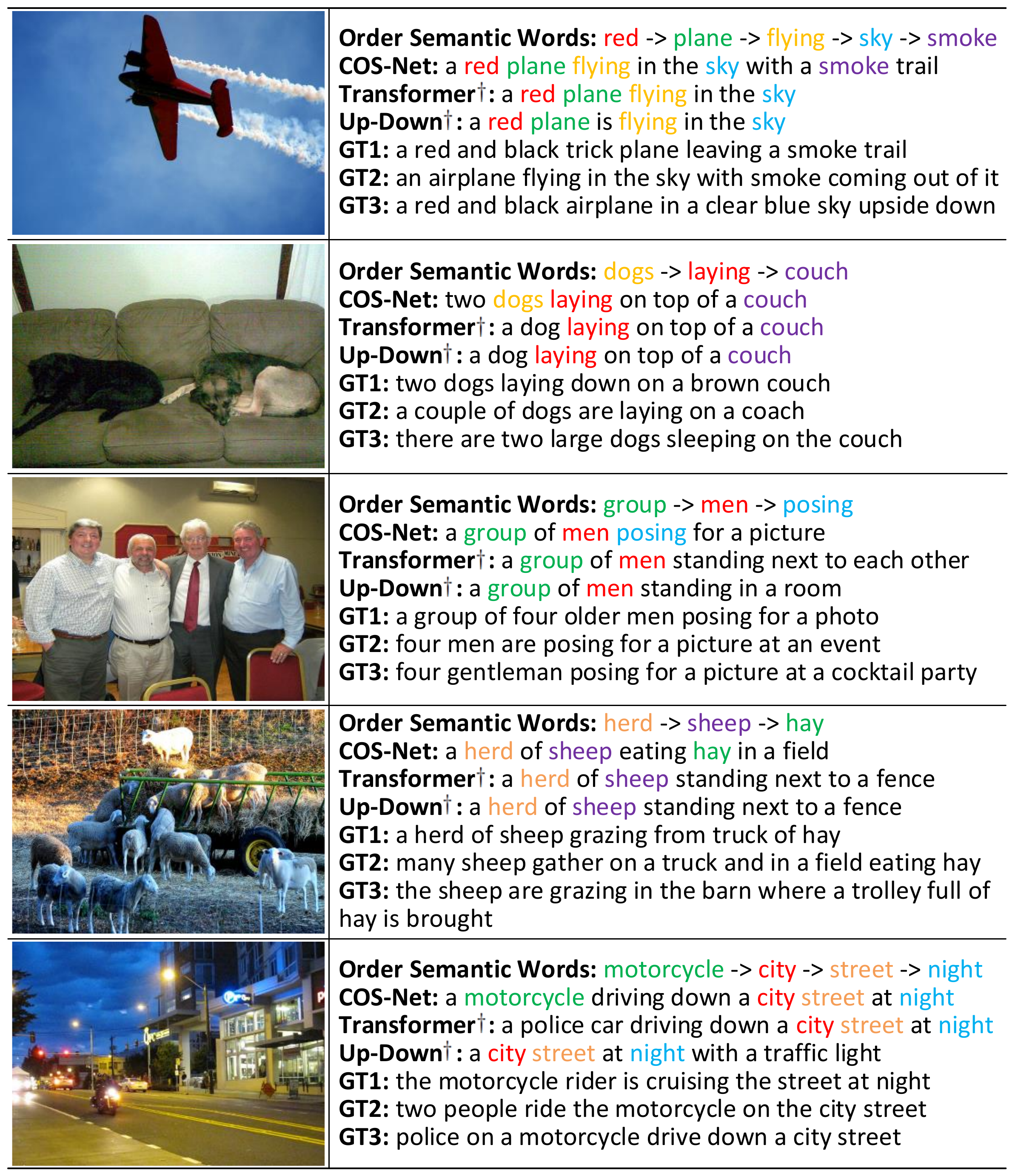}}
\vspace{-0.13in}
\caption{\small Qualitative results of our COS-Net, Transformer \textdagger~and Up-Down \textdagger~, coupled with ground-truth descriptions (GT).}
\label{fig:result}
\vspace{-0.28in}
\end{figure}

In order to qualitatively show the effectiveness of COS-Net, we showcase several qualitative results of our COS-Net and two upgraded baselines (i.e., Transformer \textdagger~and Up-Down \textdagger), coupled with the human-annotated ground-truth sentences (GT) in Figure \ref{fig:result}. In general, it is easy to observe that all the three approaches are able to produce linguistically coherent descriptions. Nevertheless, when examining the semantic relevance between visual content and generated sentence, our COS-Net outperforms the other two baselines by capturing more relevant semantic words that are worthy of mention.
For instance, in the first example, both Transformer \textdagger~and Up-Down \textdagger~only partially mine the major semantic words (red, plane, flying, and sky), while ignoring the salient semantic of smoke. Instead, COS-Net manages to comprehend all major semantics in this image (red, plane, flying, sky, and smoke) and further allocates them in linguistic order as humans, yielding both visually-grounded and linguistically coherent description.

\section{Conclusion and Discussion}

In this work, we delve into the idea of comprehending and ordering the rich semantics in an image for image captioning. To verify our claim, we present a new Transformer-style encoder-decoder structure, i.e., COS-Net, that unifies the two processes of enriched semantic comprehending and learnable semantic ordering into a single architecture. Particularly, a CLIP-based cross-modal retrieval model is initially utilized to accumulate the primary semantic cues implied in the searched semantically similar sentences. After that, a semantic comprehender filters out the irrelevant semantic words in primary semantic cues and meanwhile infers the missing relevant semantic words. Subsequently, a semantic ranker learns to estimate the linguistic position of each semantic word, leading to a sequence of ordered semantic words. The ordered semantic words serve as additional supervisory signals to guide sentence generation. We validate our proposals through extensive experiments conducted on COCO benchmark.

\textbf{Broader Impact.}
Our COS-Net is trained to produce image descriptions based on the learnt statistics of training dataset, and as such will reflect biases naturally rooted in those data, thereby resulting in negative societal impacts. Thus more future research is necessary to address this issue.

\textbf{Acknowledgments.} This work was supported by the National Key R\&D Program of China under Grant No. 2020AAA0108600.

{\small
\bibliographystyle{ieee_fullname}
\bibliography{egbib}
}

\end{document}